\documentclass[a4paper, 10pt, conference]{ieeeconf}
\usepackage[utf8]{inputenc}
\usepackage{graphicx}
\usepackage{amsmath}
\usepackage{amssymb}
\usepackage{url}
\usepackage{booktabs}
\usepackage[table,xcdraw]{xcolor}
\title{ICCV 2021 Understanding Social Behavior in Dyadic and\\ Small Group Interactions Challenge\\ \vspace{0.5cm}\textit{Fact sheet: Automatic self-reported personality recognition Track}} 

\author{\parbox{16cm}{\centering
    {\large Francisca Pessanha$^1$, Gizem Sogancioglu$^{1}$}\\
    {\normalsize
    $^1$Department of Information and Computing Sciences, Utrecht University, Utrecht, The Netherlands}}
}

\begin{document}
\maketitle
This is the fact sheet's template for the ICCV 2021 Understanding Social Behavior in Dyadic and Small Group Interactions Challenge~\cite{chalearn:iccv:2021}, ``Automatic self-reported personality recognition Track''. Please fill out the following sections carefully in a scientific writing style. Then, send the compressed project (in \texttt{.zip} format), i.e., the generated PDF, \texttt{.tex}, \texttt{.bib} and any additional files to \url{juliojj@gmail.com}, and put in the Subject of the email ``Fact Sheets: ICCV 2021 (DYAD) Challenge'', following the schedule and instructions provided in the Challenge webpage~\cite{chalearn:iccv:2021} ``\textit{Wining solutions (post-challenge), Fact Sheets}''.

%
%

%
\section{Team details}\label{teamdetails}
\begin{itemize}
    \item Team leader name: Francisca Pessanha and Gizem Sogancioglu
    \item Username on Codalab: fpessanha, gizemsogancioglu
    \item Team leader affiliation: Utrecht University
    \item Team leader email: f.pessanha@uu.nl, g.sogancioglu@uu.nl
    \item Name of other team members (and affiliation): Metehan Doyran, Heysem Kaya, Ronald Poppe, Albert Ali Salah, Almila Akdag Salah
    \item Team website URL (if any): \url{https://www.uu.nl/en/research/interaction/social-and-affective-computing/people} 
\end{itemize}
%
%

%
\section{Contribution details}

\subsection{Title of the contribution}
We propose an informed baseline to help disentangle the various contextual factors of influence in this type of case studies. For this purpose, we analysed the correlation between the given metadata and the self-assigned personality trait scores and developed a model based solely on this information. Further, we compared the performance of this informed baseline with models based on state-of-the-art visual, linguistic and audio features. For the present dataset, a model trained solely on simple metadata features (age, gender and number of sessions) proved to have superior or similar performance when compared with simple audio, linguistic or visual features based systems.

\subsection{Representative image / workflow diagram of the method}
\begin{figure}
    \centering
    \includegraphics[width=0.5\textwidth]{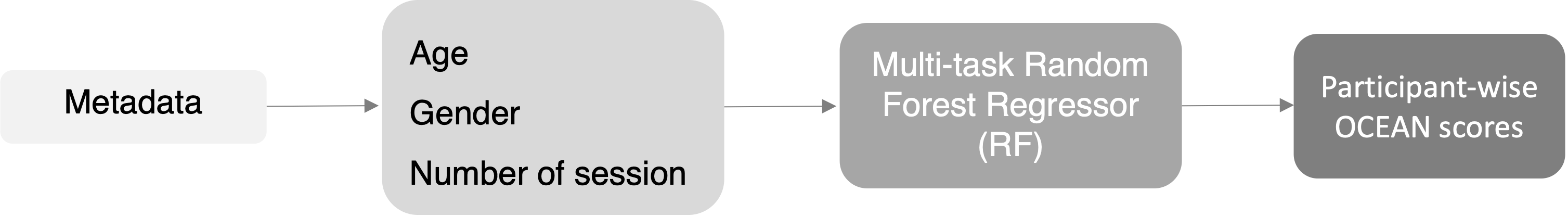}
    \caption{\textcolor{black}{Personality prediction informed baseline.}}
    \label{fig:metadata}
\end{figure}

\begin{figure*}
    \centering
    \includegraphics[width=1\textwidth]{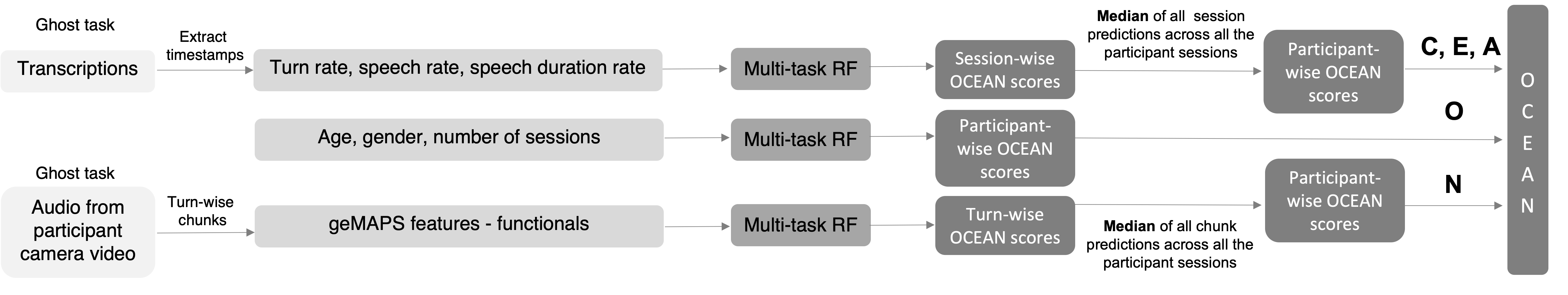}
    \caption{\textcolor{black}{Personality prediction model using linguistic, visual and metadata features. We defined the best performing models for each personality trait based on their performance on the development set and applied late fusion to get a final prediction. The best performing models for both linguistic and audio features were obtained in the ghost task.}}
    \label{fig:combined}
\end{figure*}

The pipelines for the informed (metadata based) and combined (acoustic-linguistic + metadata) systems are shown in Figures~\ref{fig:metadata} and~\ref{fig:combined}, respectively.

\subsection{Detailed method description}
The regressor used for all steps of this section was a multitask Random Forest regressor. We defined the hyperparameters of each model based on a 3-fold participant-independent cross-validation. \textcolor{black}{A model was trained for each task to evaluate what were the most representative interactions.} The best performing model was chosen based on the performance in the development set. We used both training and development sets for training in the final evaluation phase. Both per task model and fused models (using task as a feature) experimented. 

Firstly, we defined different models based on gender, age, number of sessions, background and mood dimensions. A simple model using only gender, age and number of sessions as features outperformed the development baseline in all dimensions motivating us to explore this matter further (Figure~\ref{fig:metadata}). For this purpose, we explored solutions based on linguistic, audio and video features and compared them with this informed baseline.

In the linguistic domain, we concatenated the transcriptions of the participant in the session and extracted state-of-the-art features, namely Sentence-Bert~\cite{devlin2018bert}, TFIDF~\cite{tfidf_text_mining}, LIWC~\cite{pennebaker2015development} and VAD~\cite{vad-acl2018}. Further, we handcrafted features to represent the interaction between the two participants: turn rate, speech rate and speech duration rate. In the audio domain, we sampled the audio using turn-wise chunks and extracted functionals from geMAPS for these audio segments \cite{eyben2015geneva}.  For visual, we obtained i3D features for every 2.5 seconds of the participant camera video \cite{carreira2018quo}. The final prediction for both audio and visual models was given by the median of all the predictions for the session in question. 

The best performing models in the development set where then fused as illustrated in Figure~\ref{fig:combined}. 

\subsection{Challenge results}
\label{results_test}
The best performance on the test set was achieved using the simple pipeline described in Figure~\ref{fig:metadata}.

\begin{table}[htbp]
\centering
\caption{Results from Leaderboard (Test phase) obtained by the proposed approach.}
\begin{tabular}{|c|c|c|c|c|c|c|}
\hline
\textbf{Rank position} & \textbf{O} & \textbf{C} & \textbf{E} & \textbf{A} & \textbf{N} & \textbf{MSE}\\ \hline
3 & 0.752 & 0.687 & 0.917 & 0.671 & 1.098 & 0.825 \\ \hline
\end{tabular}
\label{tab:track1:leaderboard}
\end{table}

\subsection{Final remarks}

\begin{table*}[ht]
\centering
\caption{Results obtained by the approaches represented in Figure~\ref{fig:metadata} and Figure~\ref{fig:combined} on the development set.}
\begin{tabular}{|l|c|c|c|c|c|c|}
\hline
\textbf{Method}    & \textbf{O} & \textbf{C} & \textbf{E} & \textbf{A} & \textbf{N} & \textbf{MSE}                \\ \hline
Challenge baseline & 0.6589     & 1.3541     & 1.1893     & 0.9651     & 1.1487     & 1.0632                      \\ \hline
Metadata model     & 0.6074     & 1.3285     & 1.2082     & 0.9554     & 1.1029     & 1.0405                      \\ \hline
Combined model     & 0.6074     & 1.3285     & 1.0864     & 0.8476     & 1.0955     & 0.9931 \\ \hline
\end{tabular}
\label{results_dev}
\end{table*}

\textcolor{black}{In the development phase, the combined model described in Figure \ref{fig:combined} achieved the best performance (Table \ref{results_dev}).  However, these results were not replicable in the final evaluation phase, possibly due to differences in trait balance amongst the splits and the small number of samples for both development and test sets, leading to overfitting to the development set characteristics. Therefore, the results presented in Section \ref{results_dev} were obtained with a metadata trained model.}

A correlation between personality, gender and age has been reported in previous studies. According to Marsh et al. \cite{marsh2013measurement} women have higher latent scores on all OCEAN traits, except for openness. Further, with age, individuals become happier, more self-content and self-centred and more conformed, which leads to less openness, neuroticism, extroversion and conscientiousness and more agreeability. Further, we hypothesise that, considering the time investment and social energy required for participation in a session, the number of sessions for each participant is correlated to their personality traits, namely openness. For this reason, using metadata for personality prediction is a relevant experiment.

Despite this information, the three metadata factors referred should not be sufficient for a strong personality prediction, particularly when compared with audio-visual-metadata informed systems like the challenge baseline. This fact may indicate that the dataset includes unwanted bias. Considering the small size of the dataset and the similarities in backgrounds of the participants, with a mainly highly educated population, we expect to find patterns in personality that won't generalize for a more diverse population. Further, "apparent personality'' may deviate from the "real personality'', particularly for traits low in visibility like emotional stability, which can then influence the performance of appearance-based models' performance like the baseline. \textcolor{black}{In this domain, recent work on "apparent personality'' prediction also points out the influence of inherent characteristics of the subject such as age, gender, attractiveness and ethnicity, on the way their personality is perceived \cite{principi2019effect}.}The same is true for the way the participants interact with each other, expecting different interactions when the participant is paired with similar or dissimilar personality types \cite{cuperman2009big}. So, we consider the dataset would benefit from third-person annotation to analyze the correlations between self-reported and perceived personality traits. 

In sum, our goal with the present submission was to propose an informed baseline to assess the influence of metadata in the performance of models for multimodal tasks. We argue that understandable baseline models are fundamental to access the contributions and shortcomings of complex systems in the affective computing field and contributes to the early detection of dataset related bias.

\section{Additional method details}
Please, reply if your challenge entry considered (or not) the following strategies and provide a brief explanation. For the non-binary questions, you can mark multiple options.

%
%
\begin{itemize}
\item \textbf{Mark with an X the modalities you have exploited.} (~)~Visual, (~)~Acoustic, (~)~Transcripts, (x)~Metadata, (~)~Landmark annotations, (~)~Eye-gaze vectors.\\
%
%

\item \textbf{In case you used metadata, mark with an X the types of metadata you have exploited.} (x)~Age, (x)~Gender, (~)~Country of origin, (~)~Max. level of education, (~)~Pre-session mood, (~)~Post-session mood, (~)~Pre-session fatigue, (~)~Post-session fatigue, (~)~Relationship among interactants, (~)~Task type, (~)~Task order, (~)~Task difficulty, (~)~Language, (x)~Other.\\
If ``other", or if you have used just a subset of info for a given type of metadata (e.g., just a subset of mood values), please detail: Number of sessions \\
%
%

\item \textbf{Mark with an X the tasks you used for training.} (~)~Talk, (~)~Lego, (~)~Animals, (~)~Ghost.\\
\textcolor{black}{The model proposed uses solely age, gender and number of sessions as features so, no task information is needed. In the training phase, we explored approaches using all four tasks.}
%

\item \textbf{Mark with an X the tasks you used for evaluation.} (~)~Talk, (~)~Lego, (~)~Animals, (~)~Ghost.\\
%
%

\item \textbf{Did you use the provided validation set as part of your training set?} (x) Yes, (~) No\\
If yes, please detail:\\ The final model was trained in the combined validation and training set. 
%
%

\item \textbf{Did you use any fusion strategy of modalities?} (~) Yes, (x) No\\
If yes, please detail:\\
%
%

\item \textbf{Did you use ensemble models?} (~) Yes, (x) No\\
If yes, please detail:\\
%
%

\item \textbf{Did you follow a multi-task approach or trained each trait individually?} (x) Multi-task, (~) Trained each trait individually.\\
%
%

\item \textbf{Did you use information from the other interlocutor (e.g., their visual info) to predict the personality of the target interlocutor?} (~) Yes, (x) No.\\
If yes, please detail:\\
%
%

\item \textbf{Did you use pre-trained models?} (~) Yes, (x) No\\
If yes, please detail:\\
%
%

\item \textbf{Did you use external data?} (~) Yes, (x) No\\
If yes, please detail:\\
%
%

\item \textbf{Did you use any regularization strategies/terms?} (~) Yes, (x) No\\
If yes, please detail:\\
%
%

\item \textbf{Did you use handcrafted features?} (~) Yes, (x) No\\
If yes, please detail: \\
%
%

\item \textbf{Did you use any pose estimation method?} (~) Yes, (x) No\\
If yes, please detail:\\
%
%

\item \textbf{Did you use any face / hand / body detection, alignment or segmentation strategy?} (~) Yes, (x) No\\
If yes, please detail:\\
%
%

\item \textbf{At what level of granularity did your method perform personality inference?} (~)~Frame-level, (~)~Audio/video chunk-level (i.e., short audio/video snippet), (~)~Task-level,  (~)~Session-level,  (~)~Other.\\
If ``other", please detail. If you selected ``chunk-level", please comment on the chunk length and why you selected it: Participant-level\\
%
%

\item \textbf{Did you use any aggregation method to compute a single personality prediction per participant?} (~)~Yes, (x) No\\
If yes, please detail:\\
%
%

\item \textbf{Did you use any spatio-temporal feature extraction strategy?} (~) Yes, (x) No\\
If yes, please detail:\\
%
%

\item \textbf{Did you perform any data augmentation?} \\(~) Yes, (x) No\\
If yes, please detail:\\
%
%

\item \textbf{Did you use any bias mitigation technique (e.g., rebalancing training data)?} \\(~) Yes, (x) No\\
If yes, please detail:\\
%
%

\end{itemize}

\section{Code repository}

Link to a code repository with complete and detailed instructions so that the results obtained on Codalab can be reproduced locally. This includes a list of requirements, pre-trained models, and so on. Note, training code with instructions is also required. This is recommended for all participants and mandatory for winners to claim their prize. \textbf{Organizers strongly encourage the use of docker to facilitate reproducibility}. \\

\textbf{Code repository:} \url{https://github.com/gizemsogancioglu/FGM_Utrecht}

\bibliographystyle{IEEEtran}
\bibliography{references}
\end{document}